\documentclass[sigconf]{acmart}
\usepackage{amsmath}
\usepackage{todonotes}
\usepackage{dsfont}
\usepackage{booktabs} % For formal tables
\usepackage{algorithm}
\usepackage{enumitem}
\usepackage{multirow}
\usepackage{enumerate}
\usepackage{algpseudocode}
\usepackage{subfig}
\usepackage{wrapfig}

\DeclareMathOperator*{\argmax}{arg\,max}
\newcommand{\hide}[1]{}

\newcommand{\name}{Black Box Explanations through Transparent Approximations}
\newcommand{\nameabb}{BETA}
\newenvironment{proofsketch}{%
  \renewcommand{\proofname}{Proof (Sketch)}\proof}{\endproof}

\newcommand\Fontvi{\fontsize{6}{6.2}\selectfont}
\newcommand\Fontvinew{\fontsize{7.5}{8.5}\selectfont}

\definecolor{darkgreen}{RGB}{85, 107, 47}

\newcommand{\attr}[1]{\textcolor{blue}{#1}}
\newcommand{\val}[1]{\textcolor{darkgreen}{#1}}
\newcommand{\class}[1]{\textcolor{red}{#1}}

\newcommand{\dsif}{{\bf If} }

\newcommand{\dseq}{{$=$}}

\newcommand{\dsle}{{$<$}}
\newcommand{\dsgeq}{{$\ge$} }

\newcommand{\dsand}{{\bf and} }

\newcommand{\dsthen}{{\bf then} }

\theoremstyle{plain}
\theoremstyle{definition}

\newtheorem{defn1}{Definition}

\begin{document}
\title{Interpretable \& Explorable Approximations of Black Box Models}
\author{Himabindu Lakkaraju}
\affiliation{%
  \institution{Stanford University}
}
\email{himalv@cs.stanford.edu}

\author{Ece Kamar}
\affiliation{%
  \institution{Microsoft Research}
}
\email{eckamar@microsoft.com}

\author{Rich Caruana}
\affiliation{%
  \institution{Microsoft Research}
}
\email{rcaruana@microsoft.com}

\author{Jure Leskovec}
\affiliation{%
  \institution{Stanford University}
%   \city{~}
}
\email{jure@cs.stanford.edu}
\begin{abstract}
We propose \emph{\name} (\nameabb), a novel model agnostic framework for explaining the behavior of any black-box classifier by simultaneously optimizing for fidelity to the original model and interpretability of the explanation. To this end, we develop a novel objective function which allows us to learn (with optimality guarantees), a small number of compact \emph{decision sets} each of which explains the behavior of the black box model in unambiguous, well-defined regions of feature space. Furthermore, our framework also is capable of accepting user input when generating these approximations, thus allowing users to interactively explore how the black-box model behaves in different subspaces that are of interest to the user. 
To the best of our knowledge, this is the first approach which can produce global explanations of the behavior of any given black box model through joint optimization of unambiguity, fidelity,  and interpretability, while also allowing users to explore model behavior  based on their preferences.
Experimental evaluation with real-world datasets and user studies demonstrates that our approach can generate highly compact, easy-to-understand, yet accurate approximations of various kinds of predictive models compared to state-of-the-art baselines. 
\end{abstract}

\maketitle
\section{Introduction}
The successful adoption of predictive models in settings such as criminal justice and health care hinges on how much judges and doctors can understand and trust the functionality of these machine learning models. Only if decision makers have a clear understanding of the behavior of predictive models, they can evaluate when and how much to depend on these models, detect potential biases in them, and develop strategies for further model refinement. However, the increasing complexity of predictive models is making it harder to explain or reason about their behavior~\cite{ribeiro2016should}, thus, emphasizing the need for tools which can explain the complex behavior of predictive models in a faithful and interpretable manner.  

Prior research on interpretable machine learning mainly focused on learning predictive models from scratch which were human understandable. Examples of such models include decision trees~\cite{rokach2005top}, decision lists~\cite{letham2015interpretable}, decision sets~\cite{lakkarajuinterpretable}, linear models, generalized additive models~\cite{lou2012intelligible} etc. %More recently, the increased adoption of complex machine learning models such as deep neural networks and boosted trees has given rise to a new research direction where the goal is to explain the complex behavior of black-box predictive models in a faithful and interpretable manner.
%Wei et. al.~\cite{koh2017understanding} reason about individual predictions of black box models by tracing them back to the data points in the training set which are most responsible for that prediction without providing a global explanation of the black box model behavior
More recently, Ribeiro et. al.~\cite{ribeiro2016should} and Wei et. al.~\cite{koh2017understanding} proposed approaches to explain individual predictions of any black box classifier. Ribeiro et. al.~\cite{ribeiro2016should} proposed an approach which explains individual predictions of any classifier by generating locally interpretable models. They then approximate the global behavior of the classifier by choosing certain representative instances and their corresponding locally interpretable models. This approach, however, does not clearly specify which of the multiple locally interpretable models are applicable to which part of the feature space. %In addition, the fidelity of the local models to their respective regions is also hard to quantify. Another potential approach for global explanation of black box models is to employ interpretable models such as decision sets, decision trees etc. to predict labels assigned by black box models. However, the corresponding learning algorithms are not designed to return solutions which maintain the right trade-offs between fidelity to the original model and interpretability of the explanations (number of rules, predicates, etc.)~\cite{rokach2005top}.

\begin{figure}[ht!]
\centering
%\footnotesize	
\tiny
	\begin{tabular}{|l|}
	\hline \\ \\
    \dsif \attr{Age} \dsle \val{50} \dsand \attr{Male} \dseq \val{Yes}:\\ \\
    \; \; \; \; \; \; \dsif \attr{Past-Depression} \dseq \val{Yes} \dsand \attr{Insomnia} \dseq \val{No} \dsand \attr{Melancholy} \dseq \val{No}, \dsthen \class{Healthy}\\ \\
    \; \; \; \; \; \; \dsif \attr{Past-Depression} \dseq \val{Yes} \dsand \attr{Insomnia} \dseq \val{Yes} \dsand \attr{Melancholy} \dseq \val{Yes} \dsand \attr{Tiredness} \dseq \val{Yes}, \dsthen \class{Depression}\\ \\ \\
     %\; \; \; \; \; \;  \dsif \color{magenta}{\textbf{<none-of-the-above>}},   \color{black}{\dsthen} \class{Healthy} \\ \\ \\
    \dsif \attr{Age} \dsgeq \val{50} \dsand \attr{Male} \dseq \val{No}: \\ \\ 
\; \; \; \; \; \; \dsif \attr{Family-Depression} \dseq \val{Yes} \dsand \attr{Insomnia} \dseq \val{No} \dsand \attr{Melancholy} \dseq \val{Yes} \dsand \attr{Tiredness} \dseq \val{Yes}, \dsthen \class{Depression} \\ \\ 
\; \; \; \; \; \; \dsif \attr{Family-Depression} \dseq \val{No} \dsand \attr{Insomnia} \dseq \val{No} \dsand \attr{Melancholy} \dseq \val{No} \dsand \attr{Tiredness} \dseq \val{No}, \dsthen \class{Healthy} \\ \\ 
\\
    \textbf{Default}: \\ \\
    \; \; \; \; \; \; \dsif \attr{Past-Depression} \dseq \val{Yes} \dsand \attr{Tiredness} \dseq \val{No} \dsand \attr{Exercise} \dseq \val{No} \dsand \attr{Insomnia} \dseq \val{Yes}, \dsthen \class{Depression} \\ \\
    \; \; \; \; \; \; \dsif \attr{Past-Depression} \dseq \val{No} \dsand \attr{Weight-Gain} \dseq \val{Yes} \dsand \attr{Tiredness} \dseq \val{Yes} \dsand \attr{Melancholy} \dseq \val{Yes}, \dsthen \class{Depression} \\ \\
\; \; \; \; \; \;\dsif \attr{Family-Depression} \dseq \val{Yes} \dsand \attr{Insomnia} \dseq \val{Yes} \dsand \attr{Melancholy} \dseq \val{Yes} \dsand \attr{Tiredness} \dseq \val{Yes}, \dsthen \class{Depression} \\ \\
        \hline
	\end{tabular}
    \caption{Explanations generated by our approach on depression dataset when approximating a deep neural network}
    \label{tab:approx}
    \end{figure}

Here, we study the problem of constructing \textit{global explanations} of black box classifiers. Our goal is to explain the behavior of any given black-box classifier as a whole (i.e., globally) instead of just reasoning about its individual predictions. 
%We propose a novel framework, \emph{\nameabb}, which constructs approximations by accounting for fidelity to the original black box model in addition to ensuring unambiguity (a clear specification of what rule to apply to each part of feature space) and interpretability of the explanations. 
To this end, we propose a framework \nameabb which constructs a small number of compact decision sets (sets of if-then rules) each of which captures the behavior of the given black box model in certain parts of the feature space (see Figure~\ref{tab:approx}). To ensure that the resulting explanations are faithful to the original model, we choose approximations based on how well they mimic the original model in terms of assigning class labels to instances. Our framework also unambiguously specifies the rationale used for assigning labels to instances in any part of the feature space by ensuring that each decision set and the corresponding decision rules explain non-overlapping parts of the feature space.  
To ensure that the resulting explanations are interpretable, we not only employ an intuitive rule based representation but also focus on minimizing its complexity in terms of the number of rules, predicates etc. 
Our framework also allows users to explore how the original model behaves in subspaces characterized by different values of the features that are of interest to the user. % To the best of our knowledge, this is the first approach which combines the notions of fidelity, unambiguity, interpretability, and interactivity to explain the behavior of any given black box model.

To address the problem at hand, we propose a novel optimization problem which incorporates all the aforementioned aspects.
While exactly optimizing our objective is an NP-hard problem, it has a specific structure which allows for provably near-optimal solutions. In particular, we prove that our optimization problem is a non-normal, non-monotone submodular function with matroid constraints. We then employ an efficient optimization procedure based on approximate local search~\cite{lee2009non} which provides the best known approximation guarantees ($\sim$ 1/5) to solve our optimization problem.
Experimental results on a real-world depression diagnosis dataset indicate that our approach can generate much less complex and high fidelity approximations compared to state-of-the-art baselines. We also carried out user studies in which we asked human subjects to reason about a black box model's behavior using the approximations generated by our approach and other state-of-the-art baselines. Results of this study demonstrate that the approximations generated by our approach allow humans to accurately and quickly reason about the behavior of complex predictive models.
\section{Our Framework}\label{sec:approach}
In this work, the goal of creating approximations which can meaningfully explain the behavior of any black box model is guided by the following properties: \\

\textbf{Fidelity:} The approximation should correctly capture the black box model behavior in all parts of the feature space. While different notions of fidelity can be defined, one possible way this can be achieved is through the labels assigned by the approximation matching the labels assigned by the black box model for most instances (ideally all instances) in the data.\\

\textbf{Unambiguity:}
The approximation should provide a single, deterministic rationale for explaining the prediction of every instance in the data and consequently should unambiguously specify the rationale used for assigning labels to instances in any part of the feature space. \\

\textbf{Interpretability:}
The approximation that we construct should be human-understandable. While choosing an interpretable representation
(e.g., rule based models, linear models, decision trees/sets) is a minimal requirement, it is not sufficient to ensure interpretability. Cognitive limitations of humans place restrictions on the complexity of the approximations that are understandable to humans. For example, a decision tree with a hundred levels cannot be considered interpretable. Therefore, it is important to not only have an intuitive representation but also to have smaller complexity (e.g., fewer rules in case of rule based models, fewer features with non-zero coefficients in case of linear models).\\

\textbf{Interactivity}
Users might want to understand the decision logic in subspaces characterized by certain feature values  (e.g., How does the model behave for patients over the age of 50 vs. patients under the age of 30?). In this case, a generic explanation of the behavior of the black box model may not be ideal -- the features the user is interested in may not even appear in this generic explanation. This scenario highlights the need for customized approximations which allow users to explore the behavior of black box models based on their preferences. 

\subsection{Our Representation: Two Level Decision Sets}
We choose two level decision sets as the representation of our approximations. The basic building block of this structure is a decision set which is a set of if-then rules that are unordered. The two level decision set can be regarded as a set of multiple decision sets, each of which is embedded within an outer if-then structure, such that the inner if-then rules represent the decision logic employed by the black box model while labeling instances within the subspace characterized by the conditions in the outer if-then clauses. Consequently, we refer to the conditions in the outer if-then rules as \emph{neighborhood descriptors} and  the inner if-then rules as \emph{decision logic rules}. 

While the expressive power of two level decision sets is the same as that of other rule based models (e.g., decision sets\textbackslash lists\textbackslash trees), the nesting of if-then clauses in a two level decision set representation enables the optimization algorithm (discussed later) to select \emph{neighborhod descriptors} and \emph{decision logic rules} such that higher fidelity can be obtained with minimal complexity thus resulting in more compact approximations compared to conventional decision sets (more details in experiments section). In addition, two level decision set representation does not have the pitfalls associated with decision lists where understanding a particular rule requires reasoning about all the previously encountered rules because of the if-else-if construct~\cite{lakkarajuinterpretable}. 
%Below we formally define two level decision sets. 

\begin{defn1}
A {\bf two level decision set} $\mathcal{R}$ is a set of rules $\{ (q_1, s_1, c_1) \cdots$ $(q_M, s_M, c_M)\}$ where $q_i$ and $s_i$ are conjunctions of $predicates$ of the form $(feature, operator, value)$ (eg., $age \geq 50$) and $c_i$ is a class label. $q_i$ corresponds to the subspace descriptor and $(s_i,c_i)$ together represent the inner if-then rules (decision logic rules) with $s_i$ denoting the condition and $c_i$ denoting the class label. A two level decision set assigns a label to an instance $\boldsymbol{x}$ as follows: if $\boldsymbol{x}$ satisfies exactly one of the rules $i$ i.e., $\boldsymbol{x}$ satisfies $q_i \wedge s_i$, then its label is the corresponding class label $c_i$. If $\boldsymbol{x}$ satisfies none of the rules in $\mathcal{R}$, then its label is assigned using a default function and if $\boldsymbol{x}$ satisfies more than one rule in $\mathcal{R}$ then its label is assigned using a tie-breaking function. \footnote{Note that the optimization problem that we formulate in Section~\ref{sec:optprob} will ensure that the need to invoke default or tie-breaking functions is minimized.}
\end{defn1}

%The default and tie-breaking functions can be defined by the user. 
In our experiments, we employ a default function which computes the majority class label (assigned by the black box model) of all the instances in the training data which do not satisfy any rule in $\mathcal{R}$ and assigns them to this majority label. For each instance which is assigned to more than one rule in $\mathcal{R}$, we break ties by choosing the rule which has a higher agreement rate with the black box model. Other forms of default and tie-breaking functions can be easily incorporated into our framework.  

\subsection{\name} 
Next, we show how to quantify the desiderata presented earlier in the context of two-level decision sets, then formulate it as an objective function and propose an optimization procedure. 
%We start by formulazing the properties of fidelity, unambiguity, and interpretability w.r.t our two level decision set representation. Next, we state the optimization problem for learning explanations of any black box model based on these properties and also discuss how interactivity can be incorporated into the objective function. We then show that exactly optimizing this formulation is NP-hard and discuss how we exploit specific characteristics of our formulation to solve the problem with optimality guarantees. 

\subsubsection{Quantifying Fidelity, Unambiguity, and Interpretability}

Table \ref{quantify} shows how we can quantify the properties discussed earlier w.r.t a two level decision set approximation $\mathcal{R}$, a black box model $\mathcal{B} $, and a dataset $\mathcal{D} = \{\boldsymbol{x}_1, \boldsymbol{x}_2 \cdots \boldsymbol{x}_N\}$ where $\boldsymbol{x}_i$ captures the feature values of instance $i$. We treat the black box model $\mathcal{B}$ as a function which takes an instance $\boldsymbol{x} \in \mathcal{D}$ as input and returns a class label.\\

\textbf{Quantifying Fidelity: }
{\bf disagreement($\mathcal{R}$)} quantifies the infidelity of approximation $\mathcal{R}$ to the black box model $\mathcal{B} $ by summing up for each rule $(q, s, c)$ in $\mathcal{R}$, the number of instances which satisfy $q \wedge s$ but for which the label assigned by the black box model $\mathcal{B}$ does not match the label $c$. \\

\textbf{Quantifying Unambiguity:}
%This notion is captured jointly by {\bf ruleoverlap($\mathcal{R}$)} and {\bf cover($\mathcal{R}$)}. 
For every pair of rules $(q_i, s_i, c_i)$ and $(q_j, s_j, c_j)$ in $\mathcal{R}$ where $i \neq j$, we compute the number of instances which satisfy both $q_i \wedge s_i$ and $q_j \wedge s_j$, sum up all these counts. This sum is denoted by {\bf ruleoverlap($\mathcal{R}$)}. Furthermore, it is important that the approximation that we generate explain or \emph{cover} as much of the feature space (ideally, all of it) as possible. This notion is captured by {\bf cover($\mathcal{R}$)}, which is the number of those instances which satisfy the condition $q \wedge s$ associated with some rule $(q,s,c)$ in $\mathcal{R}$.\\

\textbf{Quantifying Interpretability} {\bf size($\mathcal{R}$)} is the number of rules (triples of the form $(q,s,c)$) in the two level decision set $\mathcal{R}$. \\{\bf maxwidth($\mathcal{R}$)} is the maximum width computed over all the elements in $\mathcal{R}$, where each element is either a condition of some decision logic rule $s$ or a neighborhood descriptor $q$.%, and $width(s)$ is the number of predicates in the condition $s$. Similarly, $width(q)$ is defined as the total number of predicates of the neighborhood descriptor $q$.  
{\bf numpreds($\mathcal{R}$)} counts the number of predicates in $\mathcal{R}$ including those appearing in both the decision logic rules and neighborhood descriptors. Note that the predicates of neighborhood descriptors are counted multiple times as a neighborhood descriptor $q$ could potentially appear alongside multiple decision logic rules. {\bf numdsets($\mathcal{R}$)} is the number of unique neighborhood descriptors (outer if clauses) in $\mathcal{R}$. 

In a two-level decision set, each neighborhood descriptor characterizes a specific region of the feature space and the corresponding inner if-then rules specify the decision logic of the black box model within that region. To make this distinction clear, we minimize the number of overlapping features. For every pair of a unique neighborhood descriptor $q$ and a decision logic rule $s$, we compute the number of features that occur in both $q$ and $s$ ($featureoverlap(q,s)$) and then sum up these counts. The resulting sum is denoted as {\bf featureoverlap($\mathcal{R}$)}.
\begin{table}
\caption{Measures for Fidelity, Interpretability and Unambiguity}
\Fontvi
\begin{tabular}{| l | l | }
\hline
Fidelity & $disagreement(\mathcal{R}) = \sum\limits_{i=1}^{M} | \{ \boldsymbol{x} \text{ } | \text{ } \boldsymbol{x} \in \mathcal{D}, \boldsymbol{x} \text{ satisfies } q_i \wedge s_i, \mathcal{B}(\boldsymbol{x}) \neq c_i \} |$ \\
\hline
\multirow{3}{*}{Unambiguity} & $ruleoverlap(\mathcal{R}) = \sum\limits_{i=1}^{M} \sum\limits_{j=1, i \neq j}^{M} overlap(q_i \wedge s_i, q_j \wedge s_j)$ \\
 & $cover(\mathcal{R}) = | \{ x \text{ } | \text{ } x \in \mathcal{D}, x \text{ satisfies } q_i \wedge s_i \text{ where } i \in \{1 \cdots M\} \} | $ \\
\hline
\multirow{12}{*}{Interpretability} & $size(\mathcal{R}$): number of rules (triples of the form $(q,s,c)$) in $\mathcal{R}$ \\
\\
& $maxwidth(\mathcal{R}) = \max\limits_{ e \in \bigcup\limits_{i=1}^{M} (q_i \cup s_i)} width(e)$ \\
& $numpreds(\mathcal{R}) = \sum\limits_{i=1}^{M} width(s_i) + width(q_i)$ \\
& $numdsets(\mathcal{R}) = |dset(\mathcal{R})| \text{ where } dset(\mathcal{R}) = \bigcup\limits_{i=1}^{M} q_i$ \\
& $featureoverlap(\mathcal{R}) = \sum\limits_{q \in dset(\mathcal{R})} \sum\limits_{i=1}^{M} featureoverlap(q, s_i)$ \\
\hline
\end{tabular}
\normalsize
\label{quantify}
\end{table}
\subsubsection{Optimization Problem}\label{sec:optprob}
We assume we are given as inputs a dataset $\mathcal{D}$, labels assigned to instances in $\mathcal{D}$ by black box model $\mathcal{B}$, a set of possible class labels $\mathcal{C}$, a candidate set of conjunctions of predicates (Eg., Age $\geq$ 50 and Gender = Female) $\mathcal{ND}$ from which we can pick the neighborhood descriptors, and another candidate set of conjunctions of predicates $\mathcal{DL}$ from which we can choose the decision logic rules. In practice, a frequent itemset mining algorithm such as apriori~\cite{agrawal1994fast} can be used to generate the candidate sets of conjunctions of predicates. Without any input from the user, both $\mathcal{ND}$ and $\mathcal{DL}$ are assigned to the same candidate set generated by Apriori. On the other hand, if the user is interested in exploring the behavior of the black box model w.r.t some features $\mathcal{U}$ (eg., exercise and smoking) $\mathcal{ND}$ is initialized to conjunctions from the candidate set comprising only of the features in $\mathcal{U}$. %so that the neighborhood descriptiors of the resulting approximation contain only the features in $\mathcal{U}$.

In order to facilitate theoretical analysis, the metrics from Section 2.2.1 are expressed in the objective function either as non-negative reward functions or constraints. To construct non-negative reward functions, penalty terms (metrics defined previously) are subtracted from their corresponding upper bound values ($\mathcal{P}_{max}$, $\mathcal{O}_{max}$, $\mathcal{O'}_{max}$, ${F}_{max}$) which  are computed with respect to  $\mathcal{ND}$ and $\mathcal{DL}$. 

\Fontvinew
\begin{align*}
f_1(\mathcal{R}) & = \mathcal{P}_{max} - numpreds(\mathcal{R}) \text{, where } \mathcal{P}_{max} = \mathcal{P}_{max} = 2 * \mathcal{W}_{max} * |\mathcal{ND}| * |\mathcal{DL}| \\
f_2(\mathcal{R}) &= \mathcal{O}_{max} - featureoverlap(\mathcal{R}) \text{, where } \mathcal{O}_{max} = \mathcal{W}_{max} * |\mathcal{ND}| * |\mathcal{DL}| \\
f_3(\mathcal{R}) &= \mathcal{O'}_{max} - ruleoverlap(\mathcal{R}) \text{, where } \mathcal{O'}_{max} = N \times (|\mathcal{ND}| * |\mathcal{DL}|)^2 \\ 
f_4(\mathcal{R}) &= cover(\mathcal{R}) \\
f_5(\mathcal{R}) &=  \mathcal{F}_{max} - disagreement(\mathcal{R}) \text{, where } \mathcal{F}_{max} = N \times |\mathcal{ND}| * |\mathcal{DL}| 
\end{align*}
\normalsize
where ${W}_{max}$ is the maximum width of any rule in either candidate sets.  The resulting optimization problem is:
\Fontvinew
\begin{eqnarray}
\argmax\limits_{\mathcal{R} \subseteq \mathcal{ND} \times \mathcal{DL} \times \mathcal{C}} \sum\limits_{i=1}^{5} \lambda_i f_i(\mathcal{R}) \\
\text{ s.t. } size(\mathcal{R}) \leq \epsilon_1 \text{, } maxwidth(\mathcal{R}) \leq \epsilon_2 \text{, } numdsets(\mathcal{R}) \leq \epsilon_3
\end{eqnarray}
\normalsize 

$\lambda_1 \cdots \lambda_5$ are non-negative weights which manage the relative influence of the terms in the objective. These can be specified by an end user or can be set using cross validation. The values of $\epsilon_1, \epsilon_2, \epsilon_3$ are application dependent and need to be set by an end user.

\begin{theorem}
The objective function in Eqn. 1 is non-normal, non-negative, non-monotone, submodular and the constraints of the optimization problem are matroids. 
\end{theorem}
\begin{proofsketch}
The objective function is non-negative: the first term in the functions $f_1, f_2, f_3, f_5$ is an upper bound on the value that can be taken by the second term ensuring non-negativity. In the case of $f_4$, the metric \emph{cover} cannot be negative as it denotes the number of instances in the data that satisfy some rule in the approximation. $f_1(\emptyset) = \mathcal{P}_{max} \neq 0$. Since one of the terms is non-normal and objective is a non-negative linear combination, the objective function is non-normal. In order to prove  the objective is non-monotone, let us consider the function $f_1$ and two approximations $A$ and $B$ such that $A \subseteq B$ i.e., $B$ has at least as many rules as $A$. Therefore, by definition of $numpreds$ metric, $numpreds(B) \geq numpreds(A)$ which implies that $f_1(A) \geq f_1(B)$. Since $f_1$ is non-monotone and so is the entire objective function. Last, the functions $f_1$ and $f_5$ are modular and the other three functions in the objective turn out to be submodular. The constraints of the optimization problem are matroids because they satisfy the following two properties: 1) empty set satisfies each of the constraints 2) If approximations $A$, $B$ such that $|A| < |B|$ satisfies the constraints, then $A \cup e$ where $e \in B - A$ also satisfies the constraints. %For instance, if an approximation has $\leq \epsilon_1$ rules, each of its subsets also has $\leq \epsilon_1$ rules including the empty set. 
\end{proofsketch}
\vspace{-0.15in}
\begin{corollary}
The optimization problem in Eqn. 1 is NP-Hard.
\end{corollary}
\begin{proofsketch}
The objective function in Eqn. 1 is submodular and maximizing a submodular function is NP-Hard ~\cite{khuller1999budgeted}.
\end{proofsketch}

While exactly solving the optimization problem in Eqn. 1 is NP-Hard, the specific properties of the problem: non-monotonicity, submodularity, non-normality, non-negativity and the accompanying matroid constraints allow for applying algorithms with provable optimality guarantees. We employ an optimization procedure based on approximate local search (see Algorithm 1) which provides the best known theoretical guarantees ($\sim$ 1/5 approximation) for this class of problems. 
\begin{algorithm}[t]
\Fontvi
\caption{Optimization Procedure \cite{lee2009non}}
\label{alg:sls}
\begin{algorithmic}[1]
\State \textbf{Input:} Objective $f$, domain $\mathcal{ND} \times  \mathcal{DL} \times \mathcal{C}$,   parameter $\delta$, number of constraints $k$
\State $V_1 =  \mathcal{ND} \times  \mathcal{DL} \times \mathcal{C} $
\For{$i \in \{1,2 \cdots k+1\}$} %\label{op0}
%\State
%\State
\Comment{Approximation local search procedure}% to find a solution $S_i \subseteq V_i$ which maximizes the objective $f$ while satisfying the $k$ constraints}
%\State 
\State $X = V_i$; $n = |X|$; $S_i = \emptyset$
\State Let $v$ be the element with the maximum value for $f$ and set $S_i = v$
\While{there exists a delete/update operation which increases the value of $S_i$ by a factor of at least $(1 + \frac{\delta}{n^4})$ }
\State \textbf{Delete Operation:} If $e \in S_i$ such that $f(S_i \backslash \{e\}) \geq (1 + \frac{\delta}{n^4}) f(S_i)$, then $S_i = S_i \backslash e$
\State
\State \textbf{Exchange Operation}  If $d \in X \backslash S_i$ and $e_j \in S_i$ (for $1 \leq j \leq k$)  such that 
\State $(S_i \backslash e_j) \cup \{d\}$ (for $1 \leq j \leq k$) satisfies all the $k$ constraints and 
\State $f(S_i \backslash \{e_1, e_2 \cdots e_k\} \cup \{d\}) \geq (1 + \frac{\delta}{n^4}) f(S_i)$, then $S_i = S_i \backslash \{e_1, e_2, \cdots e_k\} \cup \{d\}$
\EndWhile
\State $V_{i+1} = V_i \backslash S_i$ 
\EndFor
\State \Return the solution corresponding to $\max \{f(S_1), f(S_2), \cdots f(S_{k+1})\}$
\end{algorithmic}
\end{algorithm}
\section{Experimental Evaluation}
%\textcolor{blue}{Hima editing here}\\
We evaluate our framework on a \textbf{Depression diagnosis}~\cite{lakkarajuinterpretable} dataset collected by an online health records portal comprising of medical history, symptoms, and demographic information of about 33K individuals. The class label of  each individual is either \emph{depressed} or \emph{healthy}.

\paragraph{Baselines}
We benchmark the performance of our framework against the following baselines: 1) Locally interpretable model agnostic explanations (LIME)~\cite{ribeiro2016should} 2) Interpretable Decision Sets (IDS)~\cite{lakkarajuinterpretable} 3) Bayesian Decision Lists (BDL)~\cite{letham2015interpretable}. We employ IDS and BDL to approximate other black box models by training them with the labels of the black box models as the ground truth labels. We also construct the following variants: 4) LIME-DS where each local linear model in the LIME approach is replaced with a decision set 5) \nameabb-LM where we group instances in the data based on neighborhood descriptors obtained using our approach and then fit a separate linear model for each of these neighborhoods. 

\begin{figure}[H]
\centering
\subfloat[Number of Rules]{
		\label{fig:a}
        \includegraphics[scale=0.03]{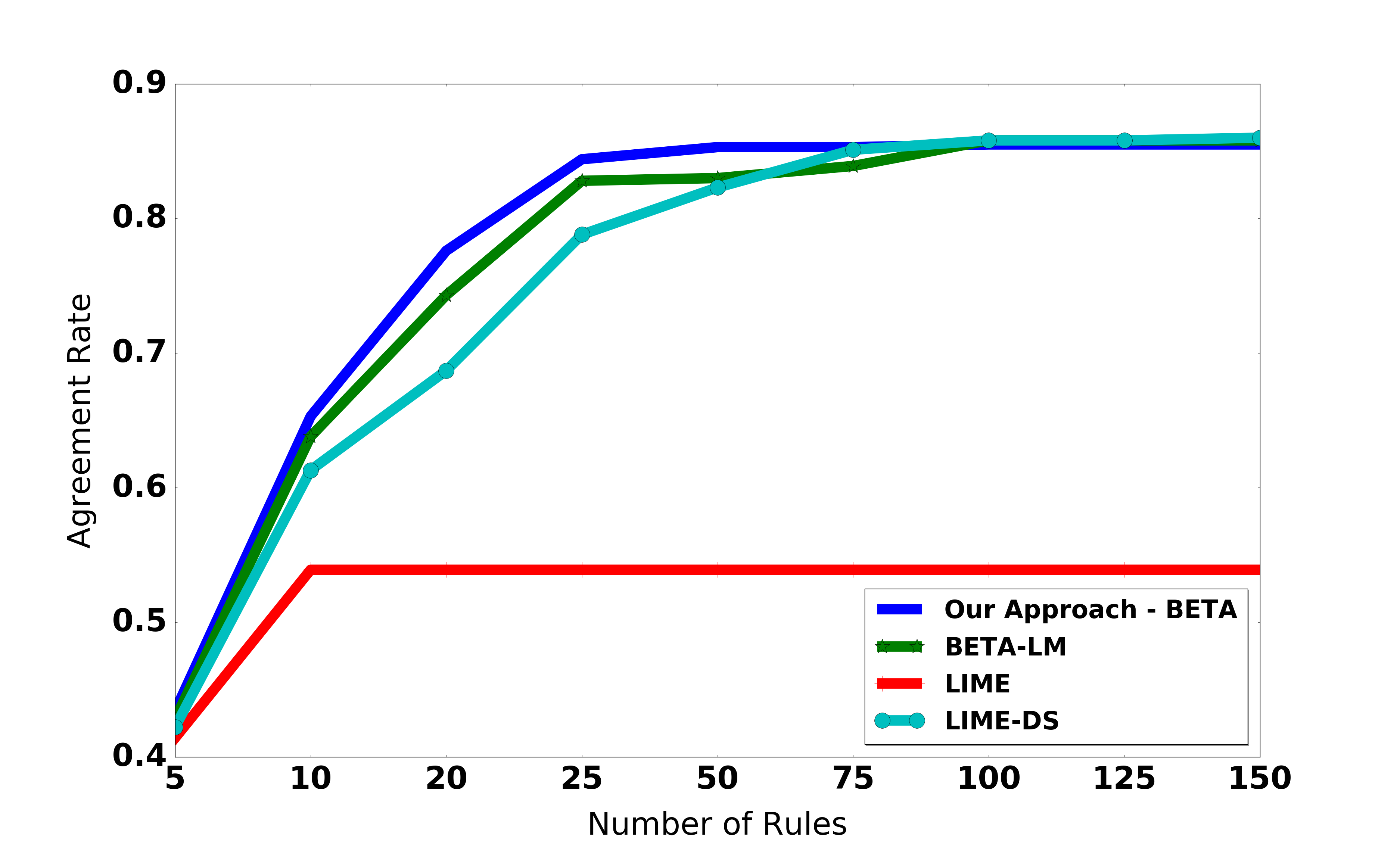}}
\subfloat[Avg. Number of Predicates]{
		\label{fig:b}
        \includegraphics[scale=0.03]{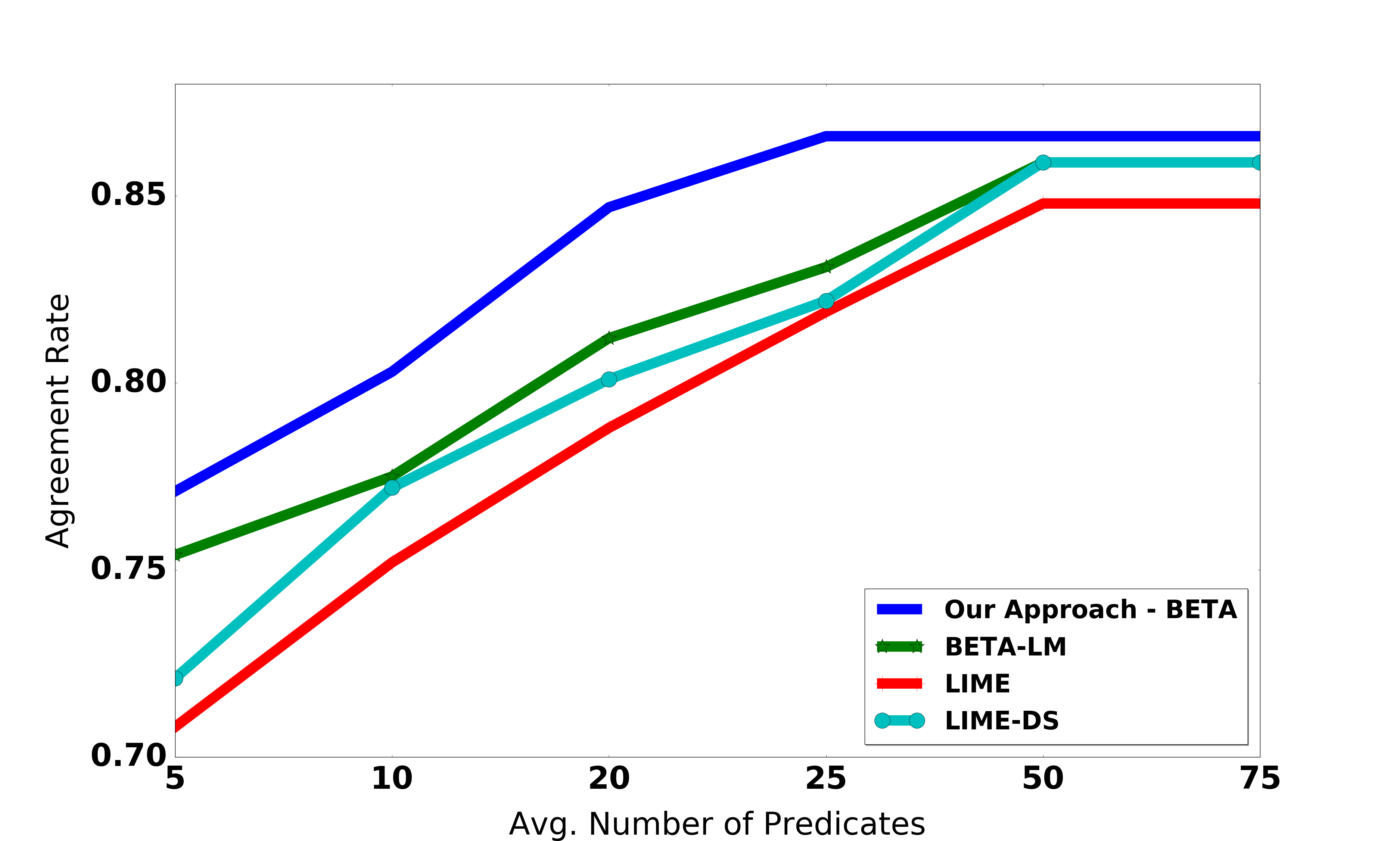}}
\subfloat[Number of Neighborhoods]{
		\label{fig:c}
        \includegraphics[scale=0.03]{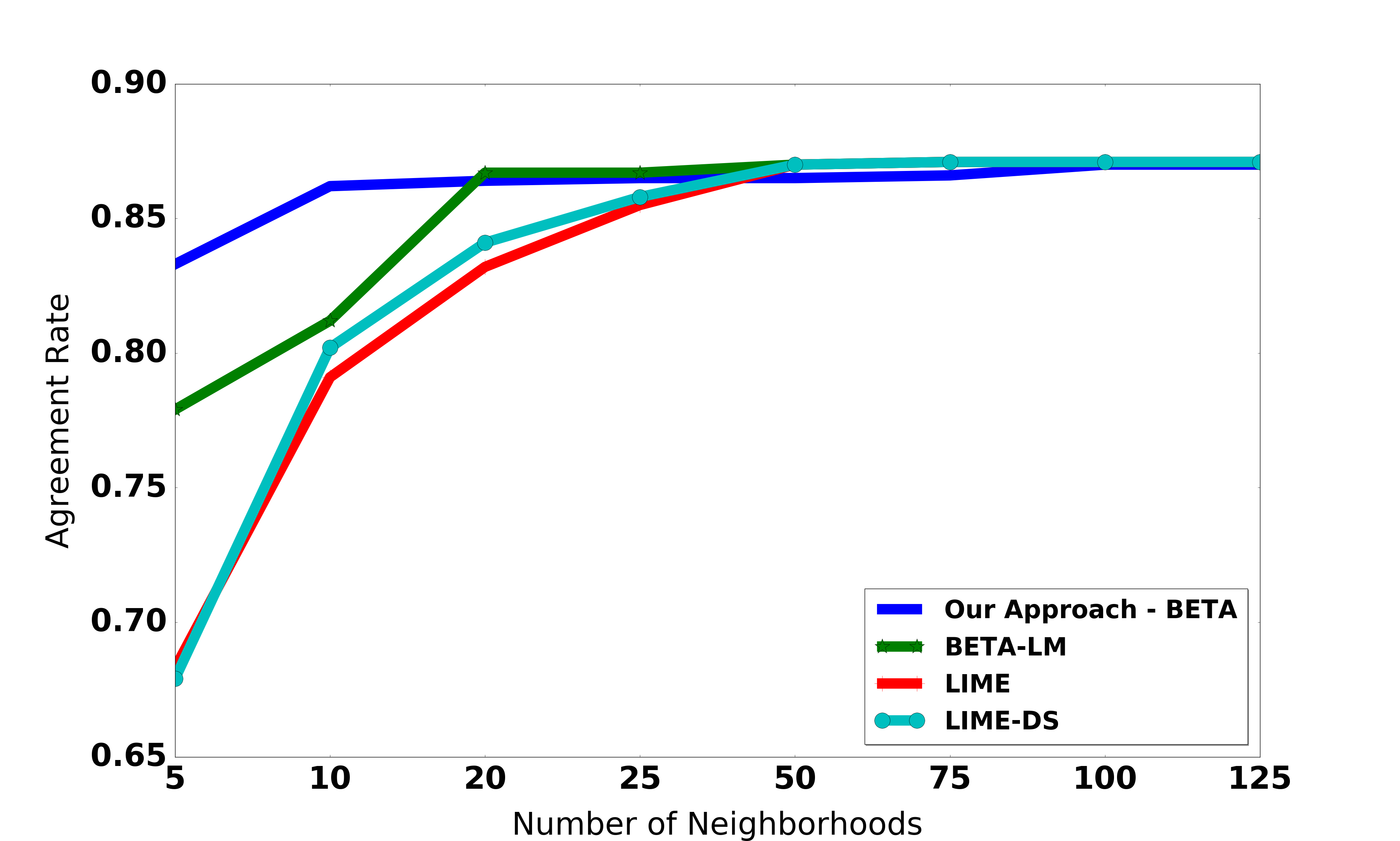}}
\caption{Fidelity vs. Interpretability Trade Offs for Depression Diagnosis Data.}
\label{fig:plots}
\end{figure}

\paragraph{Analyzing the Tradeoffs between Fidelity and Interpretability}
%no. of models vs. accuracy
%no. of rules vs. accuracy
%rule width vs. accuracy
%Fidelity and interpretability are competing objectives, where fidelity favors  details and nuances where as interpretability favors simplicity. This implies that for an approximation $A$ to be better than another approximation $B$, $A$ should rank highly on both these aspects compared to $B$. Furthermore, at different thresholds of interpretablity, the fidelity of $A$ and $B$ might differ and at certain thresholds $A$ might exhibit higher fidelity compared to $B$ and viceversa. In order to capture these nuances, we evaluate the approximations generated using our approach and other baselines via fidelity vs. interpretability curves. To this end, we define a proxy for fidelity called agreement rate which is the fraction of instances in the data for which the label assigned by the approximation is the same as that of the black box model prediction. We plot agreement rate vs. various metrics of interpretability (as outlined in Section 3) for approximations generated by our framework and other baselines:\\ \\
%Given two approximations $A$ and $B$, if $A$ achieves better performance on both of these metrics compared to $B$, then $A$ is clearly a better approximation. However, if $A$ achieves 
Fidelity and interpretability are competing objectives, where fidelity favors  details and nuances while interpretability favors simplicity. To understand how effectively different approaches trade-off fidelity with interpretability, we plot agreement rate vs. various metrics of interpretability (outlined in Section 2) for approximations generated by our framework and other baselines. We compute agreement rate, fraction of instances in the data for which the label assigned by the approximation is the same as that of the black box model prediction, as a measure of fidelity. Figures~\ref{fig:a} and~\ref{fig:b} show the plots of agreement rate vs. number of rules (\emph{size}) and agreement rate vs. average number of predicates (ratio of \emph{numpreds} to \emph{size}) for the explanations constructed to approximate a 5 layer deep neural network  using our model, LIME-DS, IDS, and BDL. Our approximations consistently demonstrate higher agreement rates at lower values of the desired metrics. For instance, at an average width of 10 predicates per rule, our approximation already reaches agreement rate of about 85\% whereas other approaches require at least 20 predicates per rule to attain this agreement rate (Figure ~\ref{fig:b}). 
%Furthermore, an approximation is considered better if it can faithfully mimic a black box model with fewer neighborhoods. 
We plot agreement rate vs. number of neighborhoods for the approximations generated by our approach and its linear variant, LIME and LIME-DS (see Figure~\ref{fig:c}). Our approximations achieve high fidelity (about 85\% agreement rate) with as few as 5 neighborhoods whereas LIME requires choosing about 20 neighborhoods to achieve the same agreement rate. %This could potentially be due to the fact that LIME greedily chooses instances which characterize neighborhoods and the resulting neighborhoods might not optimally capture the nuances of the model behavior.
%%%%%%%%%%

%We can readily evaluate the unambiguity of approximations constructed by our approach \nameabb, its linear variant \nameabb-LM, IDS, BDL using two of the metrics we outlined in Section 2, namely, \emph{ruleoverlap} and \emph{cover}. %An ideal approximation has a value of zero for \emph{ruleoverlap} and $N$ for \emph{cover} where $N$ is the number of instances in the data. 
We also found that the approximations generated using IDS and our approach also result in low values of \emph{ruleoverlap} (between $1$ and $2$\%) and high values for \emph{cover} ($95$ to $98$\%). Decision list representation by design achieves the optimal values of zero for \emph{ruleoverlap} and $N$ for \emph{cover}. % since each else-if clause ensures that every instance satisfies a single rule in the list and else clause ensures that no instance is left uncovered. 
%So, our framework and IDS are comparable on this metric where as decision lists generated using BDL are slightly better. 
%Since the notion of neighborhood is rather loosely defined (based on distances between instances) in case of LIME, these unambiguity metrics cannot be applied directly. 

\paragraph{User Studies} We designed an online user study with 33 participants, where each participant was randomly presented with the approximations (for a 5 layer deep neural network model) generated by: 1) our approach 2) IDS 3) BDL. Participants were asked $5$ questions, each of which was designed to test the user's understanding of the model behavior  in different parts of feature space. An example question is: \emph{Consider a patient who is female and aged 65 years. Based on the approximation shown above, can you be absolutely sure that this patient is Healthy? If not,  what other conditions need to hold for this patient to be labeled as Healthy?} These questions closely mimic decision making in real-world settings where decision makers would like to reason about model behavior in certain parts of the feature space. We computed the accuracy of the answers provided by users. We also recorded the time taken to answer each question and used this to computed the average time spent (in seconds) on each question. Table~\ref{wrap-tab:1} (top) show the results obtained using approximations from our model, IDS, and BDL. It can be seen that user accuracy associated with our approach was higher than that of IDS, BDL. Users were about 1.5 and 2.3 times faster when using our approximation compared to those constructed by IDS and BDL respectively. 
\begin{table}
\small
\begin{tabular}{ccc}
\\\toprule  
Approach & Human Accuracy & Avg. Time (in secs.) \\\midrule
Our Approach - BETA & 94.5\% & 160.1\\
(Non-Interactive) & & \\
IDS & 89.2\% & 231.1\\  
BDL & 83.7\% & 368.5\\  \midrule
Our Approach - BETA & 98.3\% & 78.3\\  
(Interactive) &  & \\
\bottomrule
\end{tabular}
\caption{Results of User Study.}
\label{wrap-tab:1}
\end{table}
\normalsize
We also measured the benefit obtained using \emph{interactivity}, where the approximation presented to the user is customized w.r.t to the question the user is trying to answer. For example, imagine the question above now asking about a patient who smokes and does not exercise. Whenever a user is asked this question, we showed him/her an approximation where exercise and smoking appear in the neighborhood descriptors thus simulating the effect of the user trying to interactively explore the model w.r.t these features. We recruited 11 participants for this study and we asked each of these participants the same $5$ questions as those asked in task 1. It can be seen that the time taken to answer questions is almost reduced in half compared to the setting where we showed users the same approximation each time. Answers provided are also comparatively more accurate. %, thus, demonstrating that allowing users to explore the model behavior from different perspectives can be very helpful in reasoning about its behavior in different parts of the feature space.
%\textbf{Comparing our approach with LIME (task 2)} In the final study, our goal was to carry out the comparison outlined in task 1 between our approach and LIME. However, preliminary discussions with few test subjects revealed that the loosely defined neighborhoods of LIME make it almost impossible to answer questions of the form mentioned above. We therefore carried out an online survey where we showed each participant approximations generated using our model and LIME and asked them which approximation would they prefer to use to answer questions of the form mentioned above. We recruited 12 participants for carrying out this survey and they unanimously preferred using approximations generated by our approach to reason about the model behavior. 

\bibliographystyle{abbrv}
\bibliography{sigproc} 

\end{document}